# MCTS guided Genetic Algorithm for optimization of neural network weights

Akshay Hebbar

Department of Engineering and Computer Science

Syracuse University, NewYork

ahebbar@syr.edu

*Abstract*— In this research, we investigate the possibility of applying a search strategy to genetic algorithms to explore the entire genetic tree structure. Several methods aid in performing tree searches; however, simpler algorithms such as breadth-first, depth-first, and iterative techniques are computation-heavy and often result in a long execution time. Adversarial techniques are often the preferred mechanism when performing a probabilistic search, yielding optimal results more quickly. The problem we are trying to tackle in this paper is the optimization of neural networks using genetic algorithms. Genetic algorithms (GA) form a tree of possible states and provide a mechanism for rewards via the fitness function. Monte Carlo Tree Search (MCTS) has proven to be an effective tree search strategy given states and rewards; therefore, we will combine these approaches to optimally search for the best result generated with genetic algorithms.

I. INTRODUCTION

Genetic algorithms belong to a subclass of evolutionary algorithms that provide a method for optimization based on genetic selection principles [8]. They serve a purpose in machine learning and research development, in addition to search tool optimization systems. The approach used in genetic algorithms is analogous to biological concepts of chromosome generation, with operators such as selection, crossover, mutation, and recombination. GA is a population-based approach that aims to provide a solution to successive generations. The process of evolution using GA involves starting with a random population and evolving it using crossover and mutation operators to generate children. The best-fit solution is then filtered, and the genetic process is repeated until the objective is achieved. We can observe that in the process of genetic algorithms, a tree of possible solutions is generated, and the best-fit solution is picked for successive iteration, which limits the search space and computational resources.

Genetic algorithms are used for various problem domains such as decision trees, segmentation, classification, etc. However, in this paper, our focus will be on the application of GA in optimizing neural network weights.

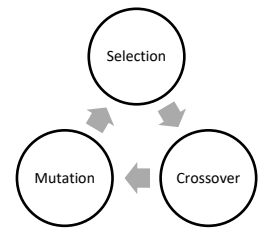

Figure 1. Simple Genetic Algorithm

The Monte Carlo Tree Search approach was developed in 2006 as an application to game-tree search. This tree search works on the principles of cumulated reward calculated from children's nodes and uses Q and N values to balance between exploration and expansion approaches. The exploration approach considers the number of nodes visited and uses a quantitative approach to discovering child nodes that have not been visited. The expansion approach follows a qualitative strategy to discovering child nodes with Q-value indicating the cumulative sum of rewards.

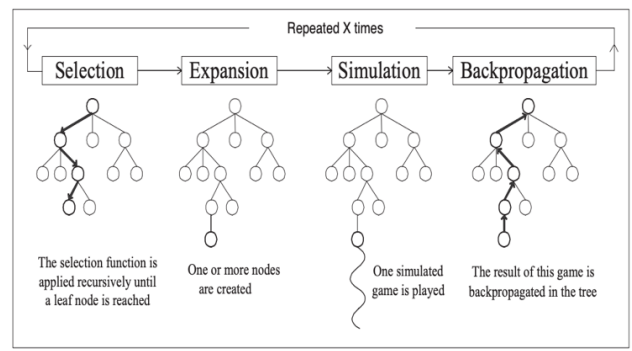

Figure 2. Outline of a Monte Carlo Tree Search

A sufficient policy for MCTS - from prior research - has been found to be UCT (upper confidence trees). UCT provides an upper confidence bound to tree search. This policy helps the search with balancing exploration vs exploitation and navigate the search space optimally. Given that the MCTS is an adversarial tree search strategy, we may be able to apply it to the entire GA tree landscape to find optimal solutions rather than exploration based on the fitness alone. Thus, we discuss the novel approach of MCTS-GA for optimization of neural network in this research.

## II. PROBLEM AND DATA DESCRIPTION

### A. GA for Neural Network weights

The first problem to consider when applying GA to optimizing the neural network weights is to confirm the validity of the child nodes generated in the process. The crossover and mutation operator applied on these weights may result in a suboptimal solution in the lower generation, but these may evolve to become better solutions in the later generations. Conversely, a solution which had highest fitness in earlier generations may end up providing a sub-optimal result in the later generation due to the nature of genetic operators used. Thus, the only way to identify the overall best solution is to let the entire tree expand and calculate fitness of each child node until a valuable solution is found. However, this solution is computation heavy, and an exhaustive tree search is rarely optimal even in cases of smaller trees. The number of nodes for a given tree which makes up for the search space is given by the below formula.

$$\frac{b^h - 1}{b - 1}, \text{ branching factor b and height h}$$

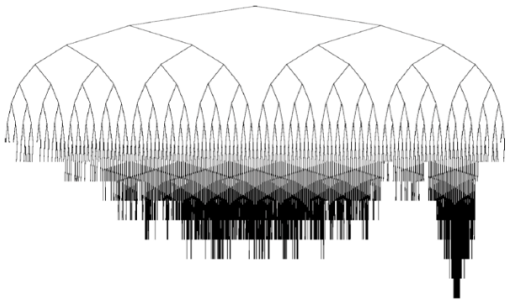

*Figure 3. Application of Monte Carlo Tree Search applied to Genetic Algorithm*

A quick calculation for a tree with branching factor of 10 and depth 10 shows the search space to be 1,111,111,111. Genetic algorithms often perform better when the size of the tree is large and with increase in tree size the number of nodes generated increases exponentially and thus the search space itself.

### B. Maintaining the Integrity of the weights

A well-known issue with genetic algorithms is the problem of competing conventions wherein the child nodes generated because of evolution are not viable and have decreased fitness. An example in case of neural network is a cross over operator applied to the weights of the neural network. The operator shuffles the weights and applies a random mutation which can propagate between layers and dimensions. The modified weights may not be appropriate for optimizing the loss function of the given neural network and may lead to an invalid solution.

### C. Data Description

The diabetes dataset is chosen to forecast the onset of diabetes mellitus. This dataset is originally from the National Institute of Diabetes and Digestive and Kidney Diseases. The objective of the dataset is to diagnostically predict whether a patient has diabetes, based on certain diagnostic measurements included in the dataset [1]. The dataset has been preprocessed by balancing using under sampling and random shuffling. A minmax scaler is used to scale the data.

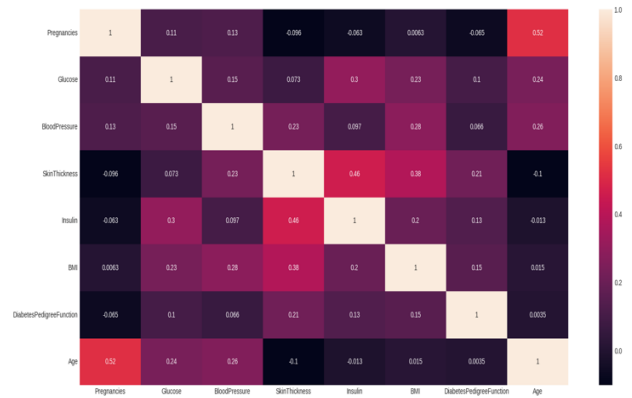

*Figure 4. Heatmap of the diabetes dataset*

For this classification problem we have developed a feedforward neural network with 4 hidden layers. There are 8 input nodes and (16-8-4-1) hidden nodes respectively in each layer.

The neural network uses sigmoid activation function, binary cross entropy loss function and Adam optimizer with 0.01 learning rate. The network is trained for 200 epochs in batch size of 10.

The weights of this neural network are used as the point of optimization for our MCTS-GA approach. Each layer weights are vectorized, combined and labelled to form an individual upon which the algorithm is applied.

### III. APPROACH

In our approach we try to address both the issues mentioned above by combining the approaches of MCTS and GA. We can take advantage of the genetic algorithm structure as it generates a tree with a mechanism to evaluate the reward in terms of the fitness of the individual. MCTS will use the same underlying tree structure along with the fitness function to calculate the Q-value which indicates the cumulative sum of rewards; the N-value which indicates the visit count of each node is also maintained for the calculation of upper bounds. The overall structure follows the complete expansion of child nodes using GA and using MCTS with UCT policy to search for the optimal nodes with fitness function as a method to assign rewards.

The process of evolution using GA involves starting with a random population which in our case is the weights of the neural network. We generate the initial population $P_0$ of certain size by adding a random uniform distribution on each dimension of the weights of the neural network and creating children with random weights.

Next, we introduce the concept of *genetic action* for MCTS as illustrated by dashed box in figure 5.

The genetic action consists of the following three genetic operations applied on the parent population.

- Genetic Action -
    - Selection
    - Crossover
    - Mutation

Selection: Various selection strategies such as roulette wheel, tournament, rank, steady state etc. are available for filtering individuals of higher fitness. The selection mechanism used in this approach is the k-way tournament selection. Tournament selection applies selection of k random individuals from the population and selects the best fit among the randomly selected individuals.

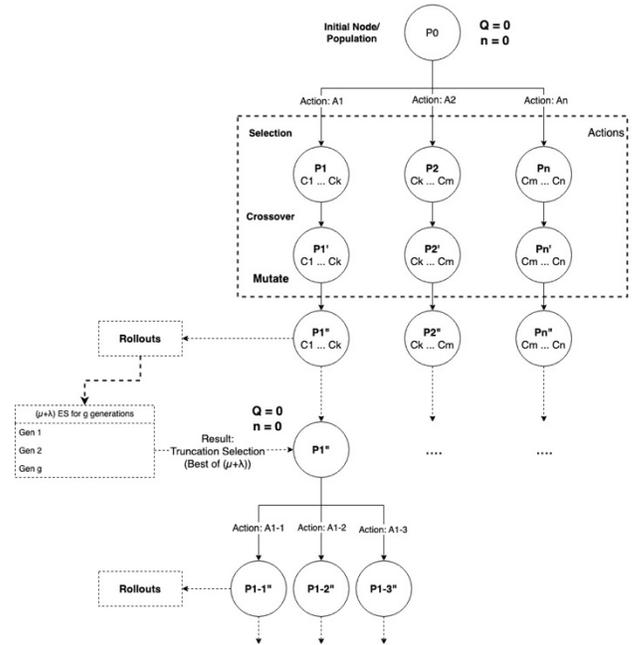

*Figure 5. Application of Monte Carlo Tree Search applied to Genetic Algorithm*

Crossover: A crossover operator is then applied on the initial population to generate children. 1-point crossover is applied where we have restricted the crossover operator to each layer of the neural network to mitigate the problem of competing conventions and maintain the integrity of the weights of the neural network.

Mutation: The mutation operator is used to introduce random mutation in the population. For our approach we have used random mutation on each layer of the neural network by swapping the weight values of 2 randomly chosen individuals. The subsequent child nodes are selected with the UCT policy of MCTS from the root node until a leaf node is found. The children among the leaf nodes are selected randomly and their corresponding Q and N values are backpropagated. The UCT policy using UCB1 is as follows.

$$UCT = Q + UCB1$$

$$UCB1 = \sqrt{\frac{c*\ln(Ni)}{N+1}}$$, $N_i$ is the $i^{th}$ child node and N is the number of child nodes

*Equation 1. Upper confidence bounds*

The next step in the process is the application of the concept of *rollout* (simulation) based on the MCTS approach. For the selected child node, an *evolutionary rollout* operation is applied wherein the individual is evolved using (μ+λ) evolutionary strategy unlike the regular random action simulations performed in prior research experiments [3]. The reason for applying the evolutionary rollout is to find out if the genetically mutated individual is of the highest fitness possible for its phenotype. We define this process as *ageing* the individual in comparison to the biological phenomenon of ageing. Thus, the ageing process introduced in rollout determines the best possible age (generation) in which the individual would be best suited to genetically mutate again. The rollout/ageing process will replace the individual if a better fitness is found in later generations of the evolutionary strategy.

This process is repeated until the specified tree depth is reached. This approach provides computational flexibility in terms of configurable parameters such as tree height, tournament selection, number of rollout generations and branching factor. These parameters can be used in combination with each other as per the computational capacity available. Thus, the application of genetic action and evolutionary rollout in amalgamation with MCTS provides the basis for the approach discussed in this paper.

## IV. RESULTS

The MCTS-GA approach is a novel mechanism for optimization and is aimed to be data agnostic. The genetic algorithm representation can be configured in multiple ways which makes this approach suitable for wide range of optimization problems. In this experiment MCTS-GA was run using UCT for 20 generations (tree depth) with rollouts configured to 10 generations and branching factor 5. The GA was run for 200 generation and the neural net was run for 200 epochs.

The results obtained from primary testing has proven to be positive. The MCTS-GA approach was able to optimize the neural network weights for better classification of the diabetes data and obtained better accuracy results.

The comparison of the results obtained to that of the original neural network and canonical genetic algorithm are shown below.

TABLE I. ACCURACY RESULTS

|   | Neural Net - SGD | Neural Net - ADAM | Genetic Algorithm | MCTS-GA |
|---|---|---|---|---|
| accuracy | 0.49 | 0.72 | 0.73 | 0.745 |
| recall | 0.42 | 0.73 | 0.77 | 0.78 |

The classification accuracy can also be noticed from the roc-auc curves indicated below.

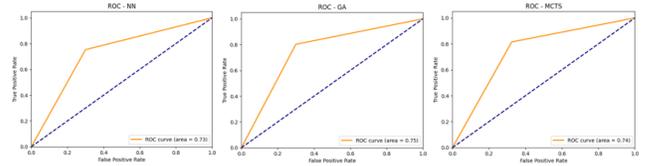

*Figure 6. ROC-AUC curve for neural network, genetic algorithm and MCTS_GA respectively*

The confusion matrix for the three approaches compared are shown below.

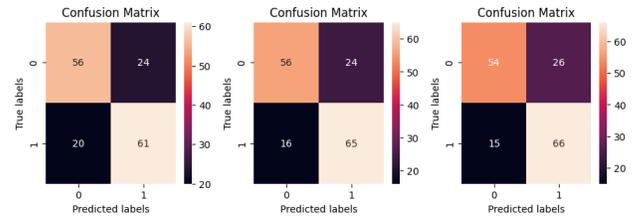

*Figure 7. confusion matrix for neural network, genetic algorithm and MCTS_GA respectively*

## V. DISCUSSION

The experiment confirms the working of MCTS-GA on optimization of neural net weights. The optimization of weights and thus the classification is seen to be better achieved by the MCTS-GA over the genetic algorithm and feedforward neural network approach. Although, the improvement is not large MCTS-GA does run in a comparable time in comparison to the other two techniques. There is scope for improvement of the algorithm discussed here and the representations of different problems are to be tested. In all, we discussed a novel approach that can prove to be a strong and valid technique for optimization techniques in the future.